\pdfoutput=1
\documentclass[11pt]{article}

\usepackage[]{acl}

\usepackage{times}
\usepackage{latexsym}

\usepackage[T1]{fontenc}
\usepackage[utf8]{inputenc}

\usepackage{microtype}
\usepackage{algorithm}
\usepackage{algorithmic}
\usepackage{booktabs}
\usepackage{amsmath}
\usepackage{multirow}
\usepackage{bbding}
\usepackage{mathrsfs}
\usepackage{subfigure}
\usepackage{amssymb}
\usepackage{graphicx} % DO NOT CHANGE THIS

\title{Generalized Intent Discovery: Learning from Open World Dialogue System}

\author{Yutao Mou$^{1*}$, Keqing He$^{2*}$, Yanan Wu$^{1}$, Pei Wang$^{1}$, Jingang Wang$^{2}$ \\
 {\bf Wei Wu$^{2}$,} {\bf Yi Huang$^{3}$,} {\bf Junlan Feng$^{3}$,} {\bf Weiran Xu$^{1}$}\thanks{\ \ The first two authors contribute equally. Weiran Xu is the corresponding author.}\\
  $^1$Beijing University of Posts and Telecommunications, Beijing, China\\
$^{2}$Meituan Group, Beijing, China\\
$^{3}$China Mobile Research Institute, Beijing, China\\
  \texttt{\{myt,yanan.wu,wangpei,xuweiran\}@bupt.edu.cn}\\
  \texttt{\{hekeqing,wangjingang,wuwei\}@meituan.com}\\
  \texttt{\{huangyi,fengjunlan\}@chinamobile.com}
  }

\begin{document}
\maketitle
\begin{abstract}
Traditional intent classification models are based on a pre-defined intent set and only recognize limited in-domain (IND) intent classes. But users may input out-of-domain (OOD) queries in a practical dialogue system. Such OOD queries can provide directions for future improvement. In this paper, we define a new task, Generalized Intent Discovery (GID), which aims to extend an IND intent classifier to an open-world intent set including IND and OOD intents.  We hope to simultaneously classify a set of labeled IND intent classes while discovering and recognizing new unlabeled OOD types incrementally. We construct three public datasets for different application scenarios and propose two kinds of frameworks, pipeline-based and end-to-end for future work. Further, We conduct exhaustive experiments and qualitative analysis to comprehend key challenges and provide new guidance for future GID research. \footnote{We release our code at \url{https://github.com/myt517/GID_benchmark}.}
\end{abstract}

\section{Introduction}

Intent classification (IC) in a dialogue system aims to identify the goal of a user query, such as \emph{BookFlight} or \emph{AddToPlaylist}. Recent neural-based models \cite{Liu2016AttentionBasedRN,Goo2018SlotGatedMF,Haihong2019ANB,Chen2019BERTFJ,He2020SyntacticGC} have achieved satisfying performance under the availability of large-scale labeled data. However, these methods face the challenge of data scarcity and poor scalability. They rely on a pre-defined intent set and supervised labels, which is limitted in some practical scenarios.

%However, these methods rely on a pre-defined intent set and supervised labels. The real online environment is dynamic, and some new intent types often appear in the input query.

% Existing intent classification models have little to offer in an open-world setting, in which many new intent categories are not defined apriori and no labeled data is available. For example, online intent classification systems are based on the pre-defined intent set, which make it only recognize limited in-domain (IND) intent categories. But plenty of input queries may be outside of the fixed intent set, which we call Out-of-Domain (OOD) intents \cite{Xu2020ADG,Zeng2021ModelingDR,Zeng2021AdversarialSL}. In recent years, OOD intent detection \cite{Hendrycks2017ABF,Larson2019AnED,Lin2019DeepUI,Ren2019LikelihoodRF,Xu2020ADG,Zheng2020OutofDomainDF} has been well studied, which aims to identify whether a user query fall outside the range of pre-defined intent set to avoid performing wrong operations. But it can't learn to distinguish the specific OOD semantic concepts. In order to further discover new OOD intents, OOD intent discovery task (also known as new intent discovery) \cite{Lin2020DiscoveringNI,Zhang2021DiscoveringNI} is proposed, which focuses on the clustering of unlabeled OOD data. However, the adopted clustering method can only group those unlabeled data into clusters, but cannot further expand the recognition scope of the in-domain intent classifier incrementally.
Existing intent classification models have little to offer in an open-world setting, in which many new intent categories are not defined apriori and no labeled data is available. These models rely on the pre-defined intent set, making it only recognize limited in-domain (IND) intent categories. But plenty of input queries may be outside of the fixed intent set, which we call Out-of-Domain (OOD) intents \cite{Xu2020ADG,Zeng2021ModelingDR,Zeng2021AdversarialSL}. In recent years, OOD intent detection \cite{Hendrycks2017ABF,Larson2019AnED,Lin2019DeepUI,Ren2019LikelihoodRF,Xu2020ADG,Zheng2020OutofDomainDF} has been well studied, which identifies whether a user query falls outside the range of pre-defined intent set to avoid performing wrong operations. But it can only safely reject OOD intents thus ignore these valuable OOD concepts for future development. Further, OOD intent discovery task (also known as new intent discovery) \cite{Lin2020DiscoveringNI,Zhang2021DiscoveringNI} is proposed to cluster unlabeled OOD data. The adopted clustering method can only group those OOD intents into clusters, but cannot further expand the recognition scope of the existing IND intent classifier incrementally.

%OOD intent discovery (also known as new intent discovery) proposed by \cite{Zhang2021DiscoveringNI} focuses on the clustering of OOD data with the aid of known IND data, which is essentially a clustering task. The adopted clustering method can only group those unlabeled data into clusters, but cannot further expand the recognition scope of the model incrementally.

%Extending existing fixed-set models to open-set intent classes is important for practical industrial dialogue systems—both to avoid performing wrong actions and to develop new skills for future improvement. But such an open-world setting means no prior OOD intent categories or supervised labels available, which makes it challenging to recognize these OOD intents incrementally.

\begin{figure}
    \centering
    \resizebox{.48\textwidth}{!}{
    \includegraphics{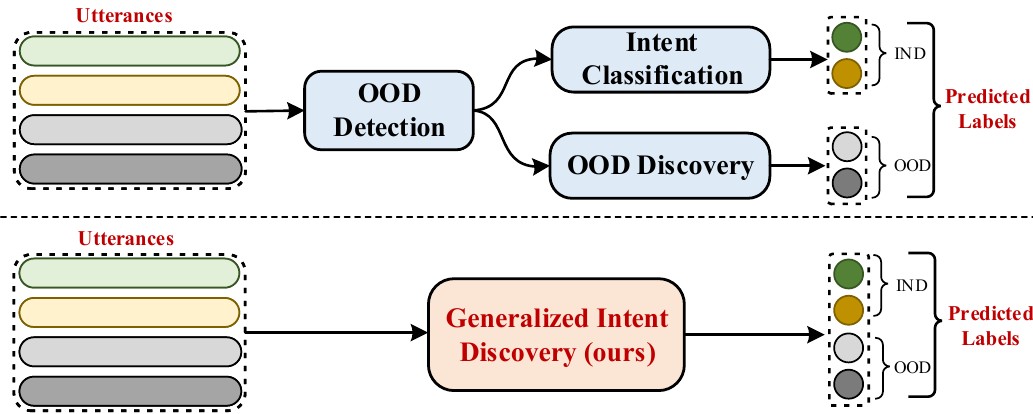}}
    \caption{Illustration of our proposed GID task. The above subfig shows a practical intent classification system where an OOD detection module firstly identifies whether a test intent belongs to OOD, then an in-domain classifier and an OOD discoverer respectively recognize IND and OOD intents. In contrast, our proposed GID can simultaneously classify a set of labeled IND intent classes and new OOD types in an end-to-end manner.}
    \vspace{-0.5cm}
    \label{fig:intro1}
\end{figure}

\begin{figure}
    \centering
    \resizebox{.48\textwidth}{!}{
    \includegraphics{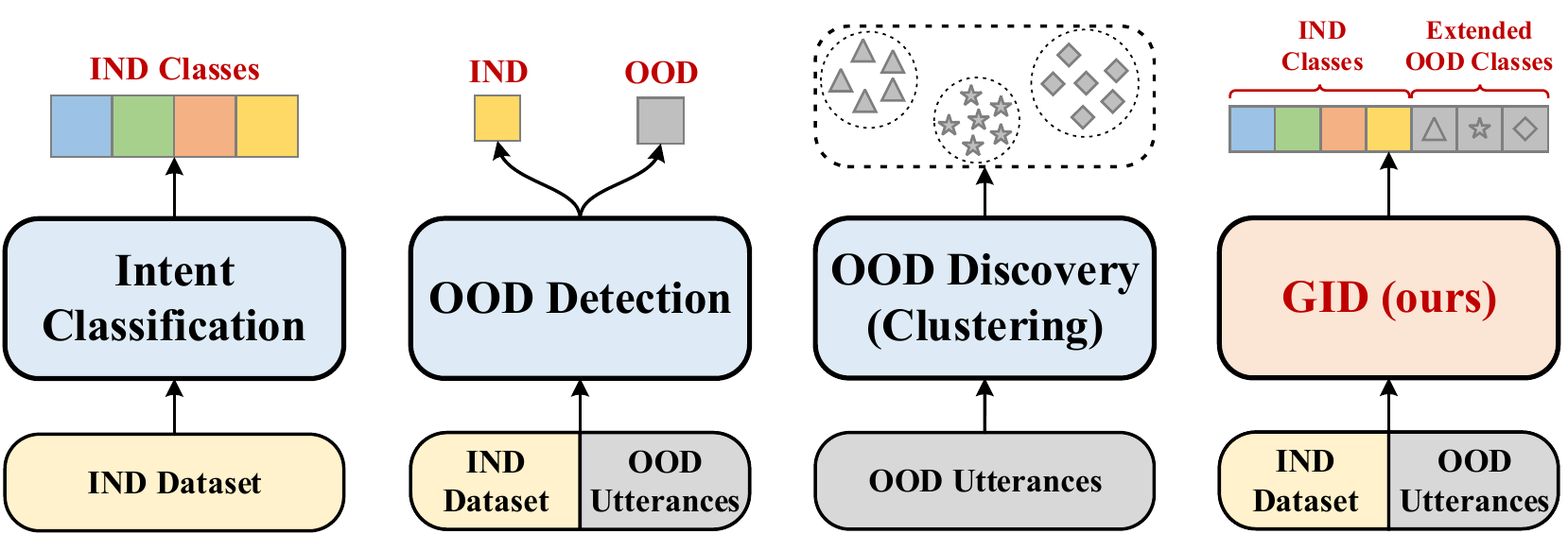}}
    \caption{The comparison of GID to other related tasks.}
    \vspace{-0.5cm}
    \label{fig:intro2}
\end{figure}

Inspired by the above issues, we introduce a new task of extending and recognizing intent categories automatically, \textbf{G}eneralized \textbf{I}ntent \textbf{D}iscovery(\textbf{GID}). GID aims to extend an existing IND intent classifier to an open-world OOD intent set, as shown in Fig \ref{fig:intro1}. The main motivation is that we hope to train a network that can simultaneously classify a set of labeled IND intent classes while discovering new ones in an unlabeled OOD set. In this way, we can enhance the capability of an IC system by expanding its recognition scope incrementally. We show a comparison of GID and existing OOD tasks in Fig \ref{fig:intro2}. Since the practical OOD intents are unsupervised, neither the OOD labels nor OOD intent schema make it different from zero-shot learning \cite{yan2020unknown, siddique2021generalized} and continual learning \cite{xu2019open} which both rely on a given label ontology, like label descriptions. Therefore, to explore unique characteristics of GID, we construct three kinds of GID benchmarks, including single domain, multiple domain, and cross-domain settings (Section \ref{datasets}). These settings denote different application scenarios which we will discuss later.

Subsequently, we propose two kinds of frameworks for GID, pipeline and end-to-end. A straightforward idea is pipeline-based methods which firstly learn OOD cluster assignments and get pseudo OOD labels, then jointly classify labeled IND data and pseudo labeled OOD data. However, pipeline-based methods separate OOD clustering and classification process, which ignores the interaction between labeled IND data and unlabeled OOD data. Besides, these pseudo OOD labels may induce severe noise to the joint classification, limiting the performance of the joint IND and OOD classifiers. Therefore, we further propose an end-to-end framework to simultaneously learn pseudo OOD cluster labels and classify IND\&OOD classes along with ground truth IND labels via a unified objective. We obtain the pseudo label of an OOD query by its augmented view in a swapped prediction way \cite{caron2020unsupervised,asano2020self,Fini2021AUO} and employ the Sinkhorn-Knopp (SK) algorithm \cite{Cuturi2013SinkhornDL} to solve the optimization problem. We leave the details to Section \ref{method}. We also perform exhaustive experiments (Section \ref{main_results}) and qualitative analysis (Section \ref{analysis}) to shed light on the challenges that current approaches face with GID. We find fine-grained OOD types, domain gap, data imbalance, real OOD noise and estimating the number of OOD types are the main challenges (Section \ref{challenges}), which provide insightful guidance for future GID work.

Our contributions are four-fold: (1) We introduce a new task, Generalized Intent Discovery (GID) which aims to extend an IND intent classifier to an open-world OOD intent set. GID helps expand the model's recognition scope and develop new skills for improving dialogue systems. (2) We construct three kinds of public GID benchmarks for different application scenarios, which help to explore the key challenges of GID comprehensively.
%and propose extensive baselines of two frameworks, pipeline-based and end-to-end for future work. 
(3) We propose an end-to-end GID framework to jointly learn clustering and classification, and extensive baselines of two frameworks, pipeline-based and end-to-end for future work. (4) We conduct exhaustive experiments and qualitative analysis to comprehend key challenges and provide new guidance for future GID research.

\section{Problem Formulation}
\label{definition}
In this section, we first briefly introduce the traditional intent classification (IC) task, then dive into the details of our proposed Generalized Intent Discovery (GID) task. 

\noindent\textbf{Intent Classification} Given a labeled in-domain (IND) dataset $\textbf{D}^{IND}=\left\{\left(x_{1}^{IND}, y_{1}^{IND}\right), \ldots,\left(x_{n}^{IND}, y_{n}^{IND}\right)\right\}$, IC aims to predict the intent class of a test query by training an IND classifier, based on the assumption that all the queries belong to a pre-defined fixed set $\mathcal{Y}^{IND}=\left\{1, \ldots, N\right\}$ of $N$ intent categories. 

\noindent\textbf{Generalized Intent Discovery} In contrast, GID is to classify queries corresponding to both labeled IND and unlabeled OOD classes. Apart from the above labeled IND dataset $\textbf{D}^{IND}$, an unlabeled OOD dataset $\textbf{D}^{OOD}=\left\{\left(x_{1}^{OOD}\right), \ldots,\left(x_{m}^{OOD}\right)\right\}$ is also given. For simplicity, we assume the number of OOD classes is specified as $M$. In practical scenarios, we can estimate the number of clusters following previous work \cite{Zhang2021DiscoveringNI} (see Section \ref{k}). Since these OOD intents are usually collected from an online IC system by rejecting low confident queries \footnote{For example, given a test query, if an IC model predicts an output with low confident probability, we can assume the query doesn't belong to any IND type but OOD intents. Please refer to related OOD detection work \cite{Xu2020ADG,Zeng2021AdversarialGD,Zheng2020OutofDomainDF} for details. In this paper, we focus on the joint classification of unlabeled OOD and labeled IND. Thus, we suppose the two sets of IND classes and OOD classes are disjoint from each other.}, the set of $N$ IND classes is assumed to be disjoint from the set of $M$ OOD classes. We also provide a discussion about real OOD noise in Section \ref{noise}. The final goal of GID is to classify an input query to the total label set $\mathcal{Y}=\left\{1, \ldots, N, N+1, \ldots, N+M\right\}$ where the first $N$ elements denote labeled IND classes and the subsequent $M$ ones denote unlabeled OOD classes. The challenges of GID come from two aspects, discovering the semantic concepts from unlabeled OOD data and jointly classifying IND\&OOD intents. On the one hand, models need to automatically cluster OOD concepts which is more difficult than supervised classification tasks. On the other hand, they require jointly recognizing IND\&OOD intents using these noisy cluster signals which may harm the final performance.

\section{Dataset}
\label{datasets}
%Since there are no existing GID datasets, 
To explore the practical significance and key challenges of GID task, we need to construct the GID dataset. However, we found that in some related tasks such as OOD intent discovery \cite{Zhang2021DiscoveringNI} and zero-shot intent detection \cite{siddique2021generalized}, the commonly used construction methods are to randomly divide the intent classification dataset into IND and OOD subset. This may not reflect real online intent classification scenarios.

We design more diverse GID dataset construction strategies, mainly in order to be able to discuss the practical significance and key challenges of GID more comprehensively. we construct three kinds of benchmark datasets GID-SD (single-domain), GID-MD (multiple-domain) and GID-CD (cross-domain) based on the two widely used intent datasets, CLINC \cite{larson2019evaluation} and Banking \cite{casanueva2020efficient}. The three settings denote different real-world application scenarios in dialogue systems. Besides, we also construct two dataset variants GID-noise and GID-imbalance to explore more severe challenges of GID tasks in real scenes. We first briefly introduce original CLINC and Banking datasets, then elaborate on GID dataset construction, and display the statistic of GID benchmarks. Finally, we introduce evaluation metrics for the GID task, accuracy and F1 score both for IND and OOD data.

\subsection{Original Intent Datasets}
CLINC contains 22,500 queries covering 150 intents across 10 domains and Banking is a fine-grained dataset in a single domain, which contains 13,083 user queries with 77 intents. We show the detailed statistics of the two original datasets in Appendix \ref{statistics}. 

\subsection{GID Dataset Construction}
% \textbf{GID Benchmarks} 
% Following the similar setting in novel visual categories discovery \cite{han2019learning, han2020automatically, zhong2021openmix, fini2021unified}, for CLINC and Banking, we keep some intent classes as OOD, and the rest is IND. To better cover the practical application scenarios of dialogue systems, we designed and constructed GID benchmark datasets for three different settings, namely single domain, multiple domain and cross domain. For the single domain scenario, we construct GID-SD by dividing the Banking dataset according to the specified OOD ratio. For the multiple domain scenario, we construct GID-MD by partitioning the CLINC dataset by intents. For the cross domain scenario, we construct GID-CD by dividing the CLINC dataset by domains. We show the construction strategies of the above three datasets in Fig\ref{}. Then we vary the number of OOD classes in the range of 20\%, 40\%, and 60\% classes to study the effect of different IND/OOD class ratios. In order to avoid the randomness of data division, we fixed three random seeds for different settings and ratios, and the final result was reported as the average of three runs. 

\textbf{GID Benchmarks} For CLINC and Banking datasets, we randomly choose the specified ratio (20\%, 40\%, 60\%) of all intent classes as OOD types, and the rest are IND, similar to \citet{Xu2020ADG,Zhang2021DiscoveringNI}.\footnote{To avoid randomness, we report the averaged experiment results of three runs for each ratio. And for each run, all the models are based on the same dataset IND/OOD split.} The original train/val/test split is fixed. We only keep IND queries with their labels and the queries belonging to OOD classes in the original train and val data. Note that GID assumes OOD training data is unlabeled so we remove OOD  queries' labels in the original train and val data. In the test set, we keep all the original IND and OOD intents and labels for evaluating metrics.\footnote{Although CLINC contains a real unlabeled OOD set, we can't use it because not able to evaluate the performance of models. We use the set for constructing a noisy GID dataset.} Considering different scenarios of dialogue systems, we construct three benchmarks, GID-SD (single-domain), GID-MD (multiple-domain) and GID-CD (cross-domain). Specifically, for the single-domain Banking dataset, we randomly select the specified ratio of all intent classes as OOD types, and the rest are IND to construct GID-SD. Since Banking has a large intent set in a single domain, we find these fine-grained OOD types are difficult to recognize (see Section \ref{GID-sd}). For the multiple-domain CLINC dataset, we propose two split strategies: (1) \textbf{Overlapping} (for GID-MD): We neglect the domain constraint and randomly split all the intent classes into the IND set and OOD set as above, which means intent categories from a domain may be divided to the two sets, which we call Domain Overlapping \footnote{Please mind IND intent classes and OOD classes are still disjoint from each other, but may belong to the same domain.}. The situation occurs where an online IC system can hardly cover all the intent classes in a domain and OOD intents may come from the same domain as IND. (2) \textbf{Non-Overlapping} (for GID-CD): We restrict IND intent classes and OOD classes are from different domains, so we select a ratio of all domains as IND and the rest as OOD. Once a domain is chosen as IND, all the intents in this domain belong to IND intent classes and vice versa. The non-overlapping setting is more practical in a real scenario where we need to transfer a business to another.

\textbf{GID Dataset Variants} To explore more severe challenges of GID tasks in real applications, we construct two variants based on GID-MD-40\%, GID-noise and GID-imbalance. (1) \textbf{GID-noise}: In the standard GID setting, we suppose the OOD data in the training set is "clean", that is, each OOD query must belong to a specific intent category. However, in practice, some OOD queries may be meaningless and not belong to any intent cluster, which we call OOD noise. We use 1350 real out-of-scope(oos) samples in CLINC, which semantically do not belong to any intent category in the training set, and add these noisy samples into the OOD train set to see if performance changes (see Section \ref{noise}). Specifically, we add different numbers of oos samples according to 5\%, 10\% and 15\% of the number of OOD samples in the training set of GID-MD-40\%. (2) \textbf{GID-imbalance}: Data imbalance is a common issue in practice. To explore the impact of OOD data imbalance, we construct imbalanced GID datasets with different imbalance ratios ($\rho=2,3,6$) by sampling each class of OOD samples in the GID-MD-40\% training set. Following \cite{zhang2021test,hong2021disentangling}, we first sort the OOD classes of GID-MD-40\% and each class is assigned an index $j (j=1,2,3,...,M)$, where $M$ denotes the total number of OOD intent categories. Then we sample from each OOD class according to $n_{j} = n_{min} \rho^{(j-1)/M}$, $j=1,2,3,...,M$, where $n_{min}$ is the least number of samples across all OOD classes. We adjust different imbalance ratios $\rho = n_{max}/n_{min}$ to simulate the degree of imbalance. $n_{max}=120$ is the max number of samples per class in GID-MD-40\%. We put the detailed statistics of GID-imbalance in Appendix \ref{imbal}.

\subsection{Statistic of GID Datasets and Evaluation}
Since different proportions of OOD intents have different statistics, here we only display the results of 40\% OOD for brevity. Table \ref{GID-40} shows the statistics of GID-SD-40\%, GID-MD-40\%, GID-CD-40\%. 

\begin{table}[t]
\centering
\resizebox{.49\textwidth}{!}{
\begin{tabular}{l|ccccccc}
\hline
Dataset & \begin{tabular}[c]{@{}c@{}}IND \\ classes\end{tabular} & \begin{tabular}[c]{@{}c@{}}OOD \\ classes\end{tabular} & \begin{tabular}[c]{@{}c@{}}IND \\ domains\end{tabular} & \begin{tabular}[c]{@{}c@{}}OOD \\ domains\end{tabular} & \begin{tabular}[c]{@{}c@{}}Train \\ samples\end{tabular}       & \begin{tabular}[c]{@{}c@{}}Val \\ samples\end{tabular}      & \begin{tabular}[c]{@{}c@{}} Test \\ samples\end{tabular}     \\ \hline
GID-SD-40\%  & 46                                                     & 31                                                     & 1                                                      & 1                                                      & 5414/3589   & 600/400  & 1840/1240 \\
GID-MD-40\% & 90 & 60 & 10 & 10 & 10,800/7200 & 1350/900 & 1350/900  \\
GID-CD-40\%  & 90                                                     & 60                                                     & 6                                                      & 4                                                      & 10,800/7200 & 1350/900 & 1350/900 \\ \hline
\end{tabular}}
\caption{Statistics of GID-SD-40\%, GID-MD-40\% and GID-CD-40\%.}
\vspace{-0.5cm}
\label{GID-40}
\end{table}

% \subsection{Evaluation Metrics}
% The traditional intent classification task uses accuracy and F1 score for evaluation, and our GID task also uses these two evaluation metrics. The difference is that our GID task needs to evaluate the performance of the methods on IND intents, OOD intents and all intents. Therefore, we need to calculate the classification accuracy and F1 score for the IND intents on the test set, and calculate the clustering accuracy\footnote{To calculate clustering accuracy, we use the Hungarian algorithm \cite{Kuhn1955TheHM} to obtain the mapping between the predicted classes and ground-truth classes.} and F1 score for the OOD intents. In addition, we need to calculate an overall accuracy and F1 score for all samples in the entire test set.

We use intent accuracy (ACC) and macro F1 as evaluation metrics for GID task. We report all IND, OOD and total (ALL) metrics where OOD and ALL ACC/F1 are the main metrics. Following \citet{Zhang2021DiscoveringNI}, we use the Hungarian algorithm \cite{Kuhn1955TheHM} to obtain the mapping between the predicted OOD classes and ground-truth classes in the test set.

\begin{figure}
    \centering
    \resizebox{0.48\textwidth}{!}{
    \includegraphics{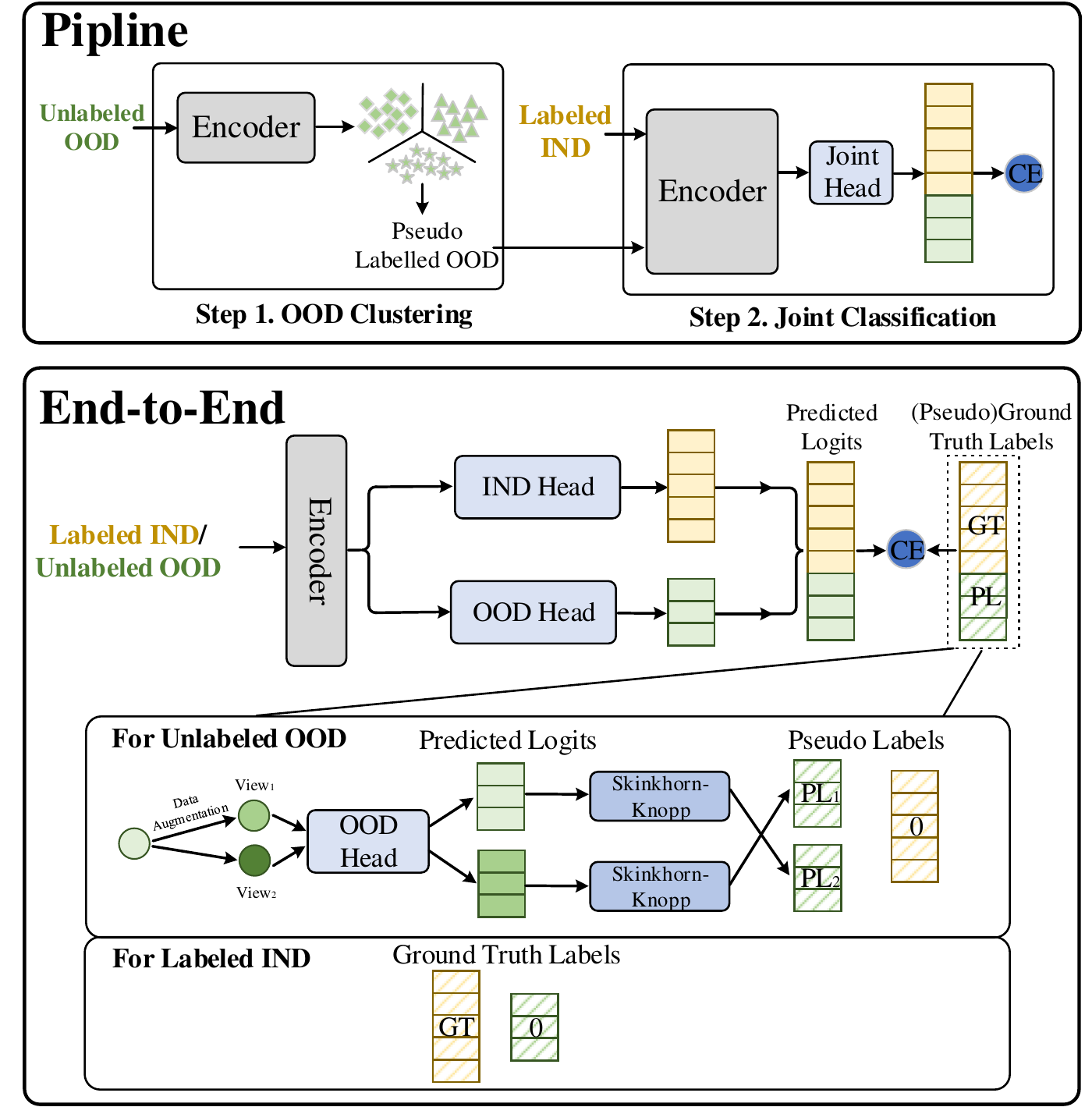}}
    \caption{Overall architecture of our proposed pipeline and end-to-end methods.}
    \vspace{-0.5cm}
    \label{fig:model}
\end{figure}

\section{Method}
\label{method}
\textbf{Overall Architecture} We extend the idea of traditional intent classification models by using pseudo OOD labels. IC calculates the $N$-dimension cross-entropy (CE) loss for labeled IND data \cite{Qin2019ASF,He2020SyntacticGC}. Similarly, we can compute (N+M)-dimension CE loss both for labeled IND and unlabeled OOD data where IND labels are given but OOD (pseudo) labels are estimated \cite{Zhang2021DiscoveringNI,Han2020AutomaticallyDA,Fini2021AUO}. Thus, the key challenge is to estimate OOD pseudo cluster labels by transferring prior IND knowledge. We propose two kinds of frameworks, pipeline and end-to-end, shown in Fig \ref{fig:model}.

% TODO：讨论下聚类伪标注的指标
\noindent\textbf{Pipeline} A simple idea is pipeline-based methods which firstly learn OOD cluster assignments, then jointly classify labeled IND data and pseudo labeled OOD data. Specifically, we use the same BERT intent encoder as DeepAligned \cite{Zhang2021DiscoveringNI} to cluster OOD data. To transfer prior knowledge, we first pre-train the encoder on IND data to get intent representations. Then, we respectively use two OOD clustering methods, k-means \cite{MacQueen1967SomeMF} and DeepAligned to obtain pseudo OOD labels $\hat{\boldsymbol{y}}^{OOD}$. Finally, we mix up all the IND and OOD data and construct the new (N+M)-dimension intent label $\boldsymbol{y}$ as follows:
{
\setlength{\abovedisplayskip}{0.1cm}
\setlength{\belowdisplayskip}{0.1cm}
\begin{equation}
\boldsymbol{y}= \begin{cases}{\left[\boldsymbol{y}^{IND}; \mathbf{0}_{M}\right]} & \mathbf{x} \in \textbf{D}^{IND} \\ {\left[\mathbf{0}_{N}; \hat{\boldsymbol{y}}^{OOD}\right]} & \mathbf{x} \in \textbf{D}^{OOD}\end{cases}
\end{equation}
}where $\boldsymbol{y}^{IND}, \hat{\boldsymbol{y}}^{OOD}$ are one-hot labels and $\mathbf{0}_{M}, \mathbf{0}_{N}$ are M or N-dimention zero vectors. We use the original CE loss to train a (N+M)-class open-set intent classifier. 

\noindent\textbf{End-to-End} The main drawback of pipeline methods is the lack of deep semantic interaction between IND and OOD data in the clustering stage, leading to poor pseudo cluster labels. To alleviate the issue, we adopt an end-to-end framework to simultaneously learn pseudo OOD cluster labels and classify IND\&OOD classes, shown in Fig \ref{fig:model}. Our motivation is that each view of an OOD intent query after data augmentation can predict the other's pseudo labels, following swapped prediction \cite{caron2020unsupervised}. And we can learn the simple pseudo-labeling process via the unified classification loss instead of extra clustering objectives. Specifically, we use the same pre-trained BERT encoder in IND data as pipeline and two independent projection layers, IND head $I$ and OOD head $O$. Given an input query, we concat the outputs of two heads as the final logit. For labeled IND intents, the ground-truth labels are easily obtained by Eq 1. We now discuss how to get the pseudo labels of unlabeled OOD intents. Inspired by \citet{caron2020unsupervised,asano2020self,Fini2021AUO}, we use the following swapped prediction way:
{
\begin{equation}
\ell_{CE}\left(\boldsymbol{x}_{1}, \hat{\boldsymbol{y}}_{2}\right)+\ell_{CE}\left(\boldsymbol{x}_{2}, \hat{\boldsymbol{y}}_{1}\right)
\end{equation}
} where $\boldsymbol{x}_{1},\boldsymbol{x}_{2}$ are two dropout-augmented \cite{Gao2021SimCSESC} views from an OOD intent query and $\hat{\boldsymbol{y}}_{1},\hat{\boldsymbol{y}}_{2}$ are corresponding pseudo labels. We use $\boldsymbol{x}_{1}$ to compute $\hat{\boldsymbol{y}}_{1}$ and $\boldsymbol{x}_{2}$ for $\hat{\boldsymbol{y}}_{2}$. A simple way of obtaining $\hat{\boldsymbol{y}}_{1}$ from $\boldsymbol{x}_{1}$ is to regard the predicted softmax logits after OOD head of $\boldsymbol{x}_{1}$ as $\hat{\boldsymbol{y}}_{1}$. But \citet{asano2020self} observes this strategy easily leads to degenerate solutions where all the intents predict the same pseudo label and are grouped into the same cluster. Therefore, we add an entropy penalty to avoid all the pseudo labels are equal to each other and keep more uniform distribution of the pseudo-labels over all the M OOD clusters. We formulate the new optimization way:
{
\begin{align}
    \hat{\mathbf{Y}}^{*} =\arg \max _{ \hat{\mathbf{Y}} \in \Gamma} \operatorname{Tr}( \hat{\mathbf{Y}} \boldsymbol{L})+\epsilon \mathrm{H}( \hat{\mathbf{Y}}) 
\end{align}
\label{eq3}
} where $\hat{\mathbf{Y}}=\left[\hat{\boldsymbol{y}}_{1}, \ldots, \hat{\boldsymbol{y}}_{B}\right]^{\top}$ is the matrix whose columns are the unknown pseudo-labels of the current batch B and $\boldsymbol{L}=\left[\boldsymbol{l}_{1}, \ldots, \boldsymbol{l}_{B}\right]$ is the predicted logits by the OOD head. $\mathrm{H}$ is the entropy function and $\epsilon$ is an hyper-parameter(we set it to 0.05 in the experiments). The goal is to obtain the best pseudo-labels $\hat{\mathbf{Y}}^{*}$ by maximizing Eq 3. And $\hat{\mathbf{Y}}$ must meet the following constraints similar to \citet{caron2020unsupervised,Fini2021AUO}:
{
\begin{align}
    \Gamma\!=\{\hat{\mathbf{Y}}\!\in\!\mathbb{R}_{+}^{M\!\times\!B}\!\mid\! \hat{\mathbf{Y}} \!\mathbf{1}_{B}\!=\!\frac{1}{M}\!\mathbf{1}_{M}\!, \!\hat{\mathbf{Y}}^{\top} \!\mathbf{1}_{M}\!=\!\frac{1}{B}\!\mathbf{1}_{B}\}
\end{align}
} where $\mathbf{1}_{B}$ denotes the vector of all ones with B dimensions. Essentially, Eq 3\&4 can be regarded as an optimal transport problem and we use the Sinkhorn-Knopp (SK) algorithm \cite{Cuturi2013SinkhornDL} to solve it.\footnote{We recommend referring to \citet{Cuturi2013SinkhornDL} for more details about the theoretical explanation of optimal transport and SK algorithm.}
After we get the pseudo OOD labels in a mini-batch, we can use Eq 1 to compute the CE loss. Note that the losses of IND and OOD data in a batch are jointly optimized. Compared to pipeline methods, our end-to-end method can simultaneously learn pseudo OOD cluster labels and distinguish IND\&OOD classes via a CE loss. Joint optimization enables semantic interaction between IND and OOD data for better knowledge transfer and to reduce noisy clustering signals. For inference, we forward the input query (including IND and OOD) to the model and obtain its prediction.

\section{Experiments and Analysis}

\begin{table*}[t]
\centering
\resizebox{0.98\textwidth}{!}{%
\begin{tabular}{l||c|cc|cc||c|cc|cc||c|cc|cc}
\hline
\multirow{3}{*}{Method} & \multicolumn{5}{c||}{GID-SD-20\%}                                               & \multicolumn{5}{c||}{GID-SD-40\%}                                               & \multicolumn{5}{c}{GID-SD-60\%}                                               \\ \cline{2-16}
                        & IND                  & \multicolumn{2}{c|}{OOD} & \multicolumn{2}{c||}{ALL} & IND                  & \multicolumn{2}{c|}{OOD} & \multicolumn{2}{c||}{ALL} & IND                  & \multicolumn{2}{c|}{OOD} & \multicolumn{2}{c}{ALL} \\
                        & ACC                  & ACC         & F1        & ACC         & F1        & ACC                  & ACC         & F1        & ACC         & F1        & ACC                  & ACC         & F1        & ACC         & F1        \\ \hline 
 k-means                          & 91.29    & 70.50    & 71.43    & 87.21    & 86.90     & 90.38    & 62.34     & 62.44    & 78.99 & 78.32    & 90.40     & 51.58    & 51.96    & 67.08    & 66.70  \\ \hline
 DeepAligned & 92.00	& 76.44	& 77.40	& 88.94	& 88.60 &91.72	&69.11	&69.72	&82.57	&82.10  &90.97	&59.55	&59.51	&72.05	&71.42                      \\ \hline
  DeepAligned-Mix  & 85.62	& 56.28	& 60.26	& 79.90	& 78.20	&82.30	&54.97	&59.79	&71.30	&69.60	&80.70	&52.66	&54.66	&63.95	&61.92                             \\ \hline
  End-to-End  &92.82	&\textbf{81.78}	&\textbf{83.53}	&\textbf{90.67}	&\textbf{90.64}	&92.84	&\textbf{72.28}	&\textbf{73.28}	&\textbf{84.49}	&\textbf{84.10}	&\textbf{92.45}	&\textbf{62.63}	&\textbf{62.65}	&\textbf{74.59}	&\textbf{73.99}                         \\ \hline
\end{tabular}
}
\caption{Performance on GID-SD (single-domain). 20\%, 40\% and 60\% denotes the ratio of OOD intents. Results are averaged over three random run.(p < 0.01 under t-test)}
\label{tab:main_result}
\end{table*}

\begin{table*}[t]
\centering
\resizebox{0.98\textwidth}{!}{%
\begin{tabular}{l||c|cc|cc||c|cc|cc||c|cc|cc}
\hline
\multirow{3}{*}{Method} & \multicolumn{5}{c||}{GID-MD-20\%}                                               & \multicolumn{5}{c||}{GID-MD-40\%}                                               & \multicolumn{5}{c}{GID-MD-60\%}                                               \\ \cline{2-16}
                        & IND                  & \multicolumn{2}{c|}{OOD} & \multicolumn{2}{c||}{ALL} & IND                  & \multicolumn{2}{c|}{OOD} & \multicolumn{2}{c||}{ALL} & IND                  & \multicolumn{2}{c|}{OOD} & \multicolumn{2}{c}{ALL} \\
                        & ACC                  & ACC         & F1        & ACC         & F1        & ACC                  & ACC         & F1        & ACC         & F1        & ACC                  & ACC         & F1        & ACC         & F1        \\ \hline 
 k-means                          &97.22	&76.22	&75.03	&93.02	&92.74	&97.26	&73.00	&72.66	&87.56	&87.08	&95.00	&65.11	&63.68	&77.02	&76.09
  \\ \hline
 DeepAligned                         &97.83	&90.89	&91.08	&96.43	&96.32	&97.85	&87.55	&87.14	&93.70	&93.29	&97.67	&83.38	&82.78	&89.10	&88.52
  \\ \hline
 DeepAligned-Mix                          &95.91	&81.93	&83.93	&93.11	&92.54	&92.86	&81.70	&83.30	&88.12	&87.42	&92.59	&78.34	&79.88	&84.05	&82.74
  \\ \hline
  End-to-End                         &98.17	&\textbf{95.26}	&\textbf{96.08}	&\textbf{97.58}	&\textbf{97.59}	&98.32	&\textbf{91.92}	&\textbf{92.46}	&\textbf{95.78}	&\textbf{95.73}	&98.26	&\textbf{87.63}	&\textbf{87.84}	&\textbf{91.88}	&\textbf{91.78}
  \\ \hline
\end{tabular}
}
\caption{Performance on GID-MD (multiple-domain).}
\label{tab:main_result}
\end{table*}

\begin{table*}[t!]
\centering
\resizebox{0.98\textwidth}{!}{%
\begin{tabular}{l||c|cc|cc||c|cc|cc||c|cc|cc}
\hline
\multirow{3}{*}{Method} & \multicolumn{5}{c||}{GID-CD-20\%}                                               & \multicolumn{5}{c||}{GID-CD-40\%}                                               & \multicolumn{5}{c}{GID-CD-60\%}                                               \\ \cline{2-16}
                        & IND                  & \multicolumn{2}{c|}{OOD} & \multicolumn{2}{c||}{ALL} & IND                  & \multicolumn{2}{c|}{OOD} & \multicolumn{2}{c||}{ALL} & IND                  & \multicolumn{2}{c|}{OOD} & \multicolumn{2}{c}{ALL} \\
                        & ACC                  & ACC         & F1        & ACC         & F1        & ACC                  & ACC         & F1        & ACC         & F1        & ACC                  & ACC         & F1        & ACC         & F1        \\ \hline 
 k-means   &97.39	&75.78	&75.79	&92.98	&92.72	&97.70	&61.67	&60.43	&83.20	&82.30	&96.44	&54.67	&53.69	&71.38	&70.57
                         \\ \hline
  DeepAligned   &97.83	&84.81	&84.22	&95.23	&95.01	&97.85	&78.55	&77.81	&90.12	&89.68	&97.33	&76.15	&74.80	&84.62	&83.60
                         \\ \hline
  DeepAligned-Mix  &97.15	&77.41	&77.7	&93.20	&92.53	&97.33	&72.41	&71.54	&87.36	&86.21	&93.89	&75.63	&74.29	&82.93	&81.37
                          \\ \hline
  End-to-End  &97.92	&\textbf{87.41}	&\textbf{87.55}	&95.81	&\textbf{95.75}	&98.00	&\textbf{79.19}	&\textbf{79.06}	&90.46	&\textbf{90.28}	&98.22	&\textbf{78.01}	&\textbf{77.48}	&\textbf{86.09}	&\textbf{85.63}
                          \\ \hline
\end{tabular}
}
\caption{Performance on GID-CD (cross-domain).}
\label{tab:main_result}
\end{table*}

\subsection{Baselines}
\noindent\textbf{k-means} A pipeline baseline, which first uses k-means \cite{MacQueen1967SomeMF} to cluster OOD data and obtains pseudo OOD labels, and then trains a new classifier together with IND data.

% 强调sota
\textbf{DeepAligned} Similar to k-means, the difference is that the clustering algorithm adopts DeepAligned \cite{Zhang2021DiscoveringNI}, which is the current state-of-the-art method for OOD discovery task.

\textbf{DeepAligned-Mix} This is an end-to-end approach where we extend DeepAligned for GID. DeepAligned is an iterative clustering method. In each iteration, it firstly uses k-means and an alignment strategy to cluster and label the OOD data and then computes the cross-entropy classification for representation learning. Our proposed DeepAligned-Mix mainly improves two points: (1) We mix up IND and OOD data together for iterative clustering, and the model is optimized with a unified cross-entropy loss; (2) In the inference stage, instead of using k-means for clustering, we use the classification head of the new classifier to make predictions.

\subsection{Main Results}
\label{main_results}

We conduct experiments on three benchmark GID datasets GID-SD, GID-MD and GID-CD with different OOD ratios, shown in Table \ref{tab:main_result}. In general, Our proposed end-to-end (E2E) method consistently outperforms all the baselines with a large margin. We analyze the results from three aspects: 

\noindent\textbf{Comparison of different methods} We see E2E significantly outperforms all the baselines under the three datasets and different OOD ratio settings. For example, E2E outperforms previous state-of-the-art DeepAligned by 3.14\%(OOD F1) and 2.57\%(ALL F1) on GID-SD-60\%, 5.06\%(OOD F1) and 3.26\%(ALL F1) on GID-MD-60\%, 2.68\%(OOD F1) and 2.03\%(ALL F1) on GID-CD-60\%. These prove that joint clustering and classification helps to perform more interaction between IND and OOD and obtain accurate pseudo OOD labels. We also observe E2E achieves slightly better IND ACC than pipeline methods, which means joint classification doesn't sacrifice IND performance while improving OOD recognition.

\label{GID-sd}
\noindent\textbf{Comparison of different datasets} To explore the effect of different practical scenarios, we compare the performance of the same method on different datasets. Results show metrics on GID-SD are the lowest, GID-CD is in the middle and GID-MD is the best for almost all the methods, which denotes the difficulty order is single-domain$>$cross-domain$>$multiple-domain. We argue GID-SD contains more fine-grained intent types in a single domain which makes it challenging to recognize OOD intents. Comparing CD and MD, IND and OOD types from the same domain makes it easier to transfer prior knowledge, so MD gets higher scores.

\noindent\textbf{Effect of different OOD ratios} We compare the results of different OOD ratios on the same dataset. We find with the increase of OOD ratio, the performance consistently drops. For example, E2E achieves 95.26\% OOD ACC on GID-MD-20\%, but OOD ACC decreases by 3.34\% on GID-MD-40\% and 7.63\% on GID-MD-60\%. Intuitively, the increase in the number of OOD intents makes it more difficult to distinguish them.

\subsection{Qualitative Analysis}
\label{analysis}

\begin{figure*}
    \centering
    \resizebox{\textwidth}{!}{
    \includegraphics{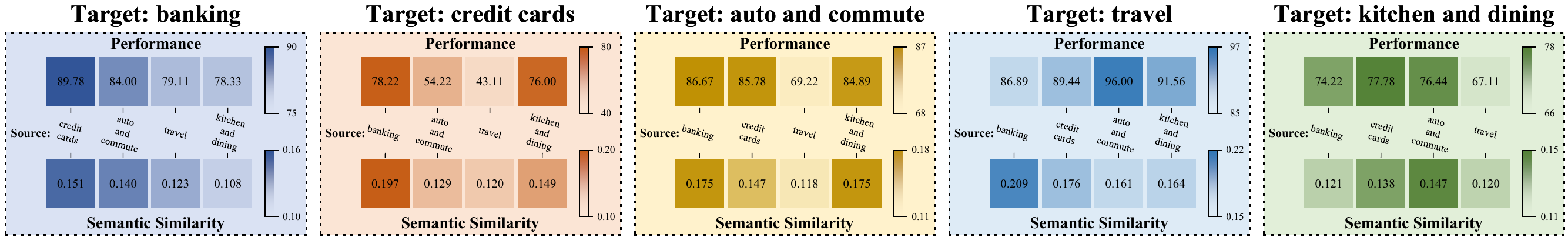}}
    % \vspace{-0.65cm}
    \caption{Cross-domain transferability from source IND to target OOD. We display OOD ACC and domain similarity scores. The larger the number is, the deeper the color is.}
    \label{fig:cross_domain_transferability}
    % \vspace{-0.5cm}
\end{figure*}

\subsubsection{Cross-Domain Transferability}
% For GID task, cross-domain transfer is an important and challenging scenario, in which we assume that IND and OOD are intent types from different domains. We selected five domains(banking, credit\_card, auto\_and\_commute, travel, kitchen\_and\_dining) and used two metrics to analyze the difficulty of transferring knowledge between different domains. First, we paired the five domains and used our end-to-end method to calculate the OOD accuracy of the GID classifier. Then we introduce a simple metric Silhouette Coefficient\cite{rousseeuw1987silhouettes} to measure transferability between domains. We first train an IND classifier, then use this classifier to extract representations of OOD samples, use pseudo labels obtained by k-means to calculate SC value. We can see that the larger the SC value, the more semantically similar the IND domain and the OOD domain are, and the greater the transferability. As shown in fig\ref{}, We compare OOD accuracy and SC value of the same OOD domain under different IND domains, and find that the more semantically similar the transfer effect is better between two domains, such as banking and credit\_card. In addition, for some domains whose semantic similarity is not obvious, the SC value can reflect the semantic similarity to a certain extent.
For GID-CD, cross-domain knowledge transfer is important and challenging. To study the effect of domain similarity on knowledge transfer, we perform a cross-domain transferability analysis in Fig \ref{fig:cross_domain_transferability}. We select five domains (banking, credit\_card, auto\_and\_commute, travel, kitchen\_and\_dining) and perform the one-to-one transfer. To measure domain similarity, we first train an IND intent classifier, then perform k-means using extracted representations of OOD samples to calculate Silhouette Coefficient (SC) values \cite{rousseeuw1987silhouettes} \footnote{Please see more details about SC in Appendix \ref{sc}.}. We can see that the larger the SC value is, the higher the similarity between IND\&OOD domains is, resulting in better OOD metrics. The results prove good cross-domain transferability comes from semantically similar domains, such as banking and credit\_card.

\begin{figure}[t]
    \flushleft
    \resizebox{0.95\linewidth}{!}{
    \includegraphics{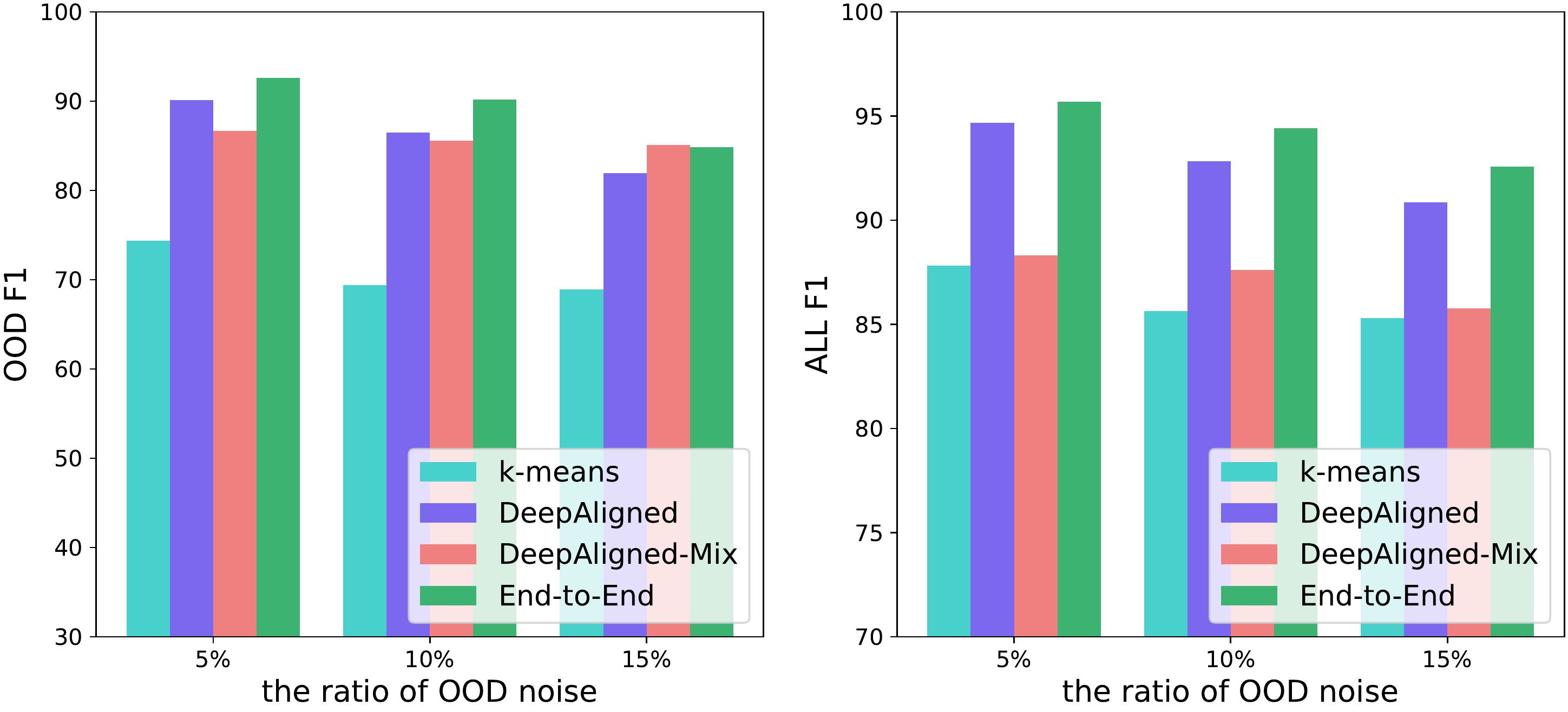}}
    % \vspace{-0.2cm}
    \caption{The impact of adding different numbers of noisy OOD samples to the training set on the performance of each GID model.}
    \label{fig:OOD_noise}
    %  \vspace{-0.65cm}
\end{figure}

\subsubsection{Effect of OOD noise}
\label{noise}
In the real world, OOD data may not necessarily belong to a certain OOD cluster, and there is often some OOD noise. We use the constructed dataset variant GID-noise to examine the impact of noisy OOD in the training set on model performance. Fig \ref{fig:OOD_noise} shows the impact of different amounts of OOD noise in the training set on model performance. The results show that as the amount of OOD noise increases, the OOD performance drops. Our proposed E2E still achieves the best performance over all baselines.
We argue that this is because the presence of OOD noise makes it difficult for the model to learn a clear cluster boundary for unlabeled OOD.

\subsubsection{Effect of imbalanced OOD data}
\label{imbalanced}
\begin{figure}[t]
    \flushleft
    \resizebox{0.98\linewidth}{!}{
    \includegraphics{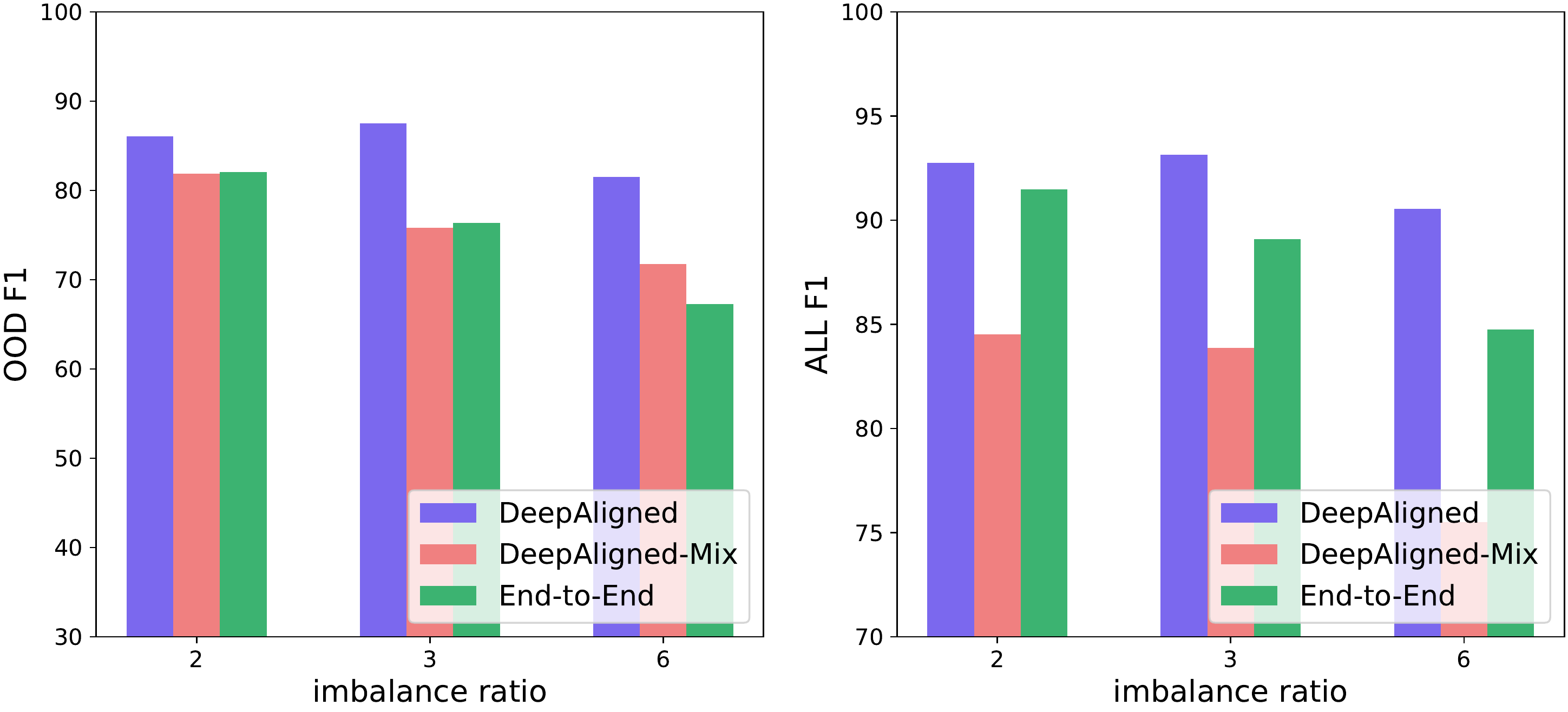}}
    % \vspace{-0.2cm}
    \caption{The impact of different imbalance ratios of OOD data on the performance of each GID model.}
    \label{fig:OOD_imbalance}
    %  \vspace{-0.3cm}
\end{figure}

Fig \ref{fig:OOD_imbalance} shows the impact of class imbalance degree of OOD data on model performance. The results show that when the imbalance degree of OOD categories increases, the performance of all models decreases significantly. We also find an interesting phenomenon that our proposed end-to-end method drops more significantly than pipeline-based DeepAligned. We argue that there are two reasons for this. (1) When our end-to-end method obtains OOD pseudo-labels, the SK algorithm is based on a strong assumption, the number of pseudo-labels for each category in a batch is uniform, which is obviously invalid in the class-imbalanced scenario. (2) E2E uses IND and OOD to jointly train the classifier. Since the number of samples in each class of IND keeps fixed to 120(equal to the number of OOD samples in the majority class of OOD), this will exacerbate the degree of imbalance and affect the accuracy of pseudo-labels for long-tail categories. Therefore, we need to further explore better pseudo-label methods in the future and how to improve the class-imbalanced defect of end-to-end methods.  

\begin{figure*}[t]
    \centering
    \subfigure[DeepAligned]{
        \includegraphics[scale=0.45]{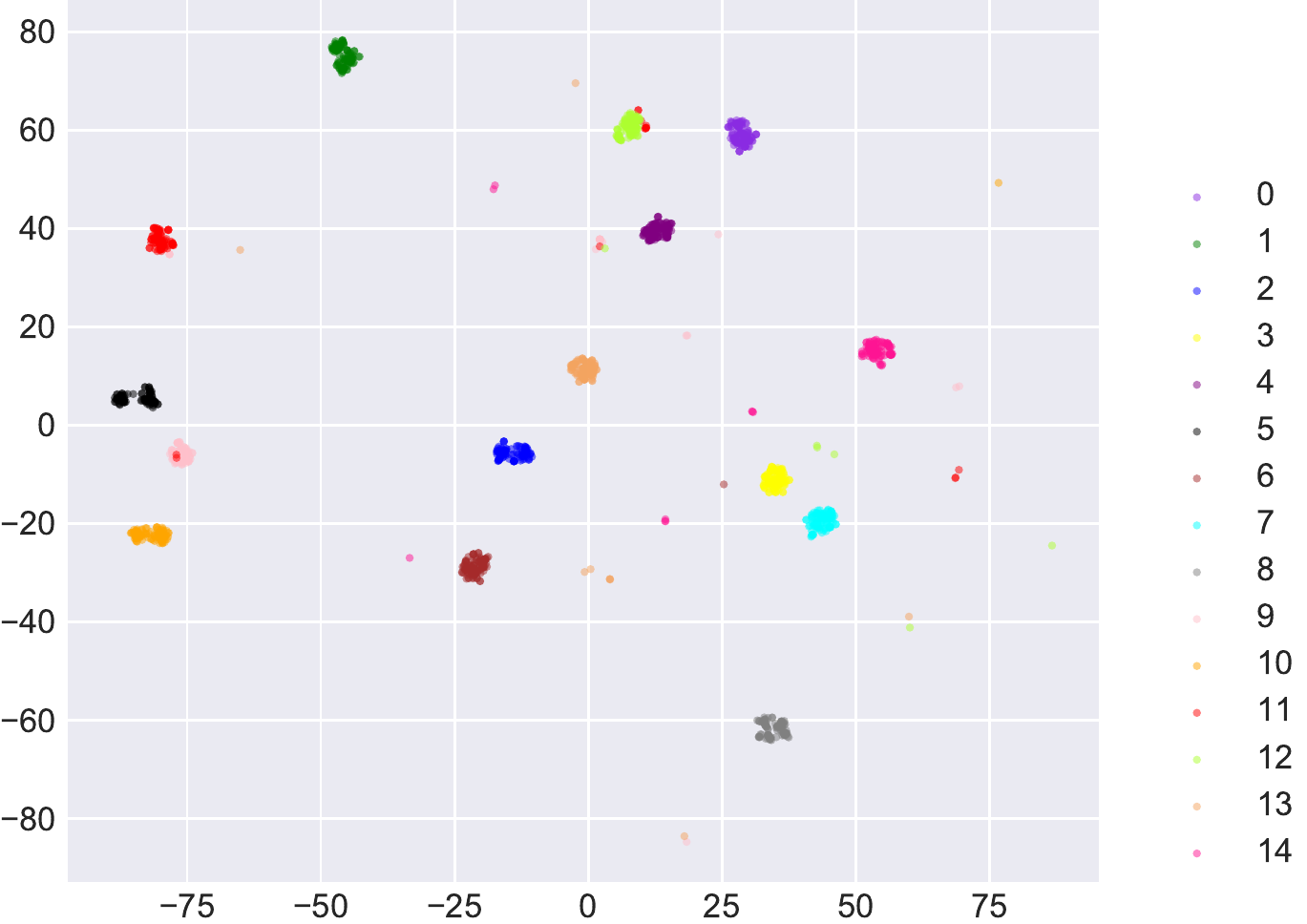}
    }
    \subfigure[End-to-End]{
        \includegraphics[scale=0.45]{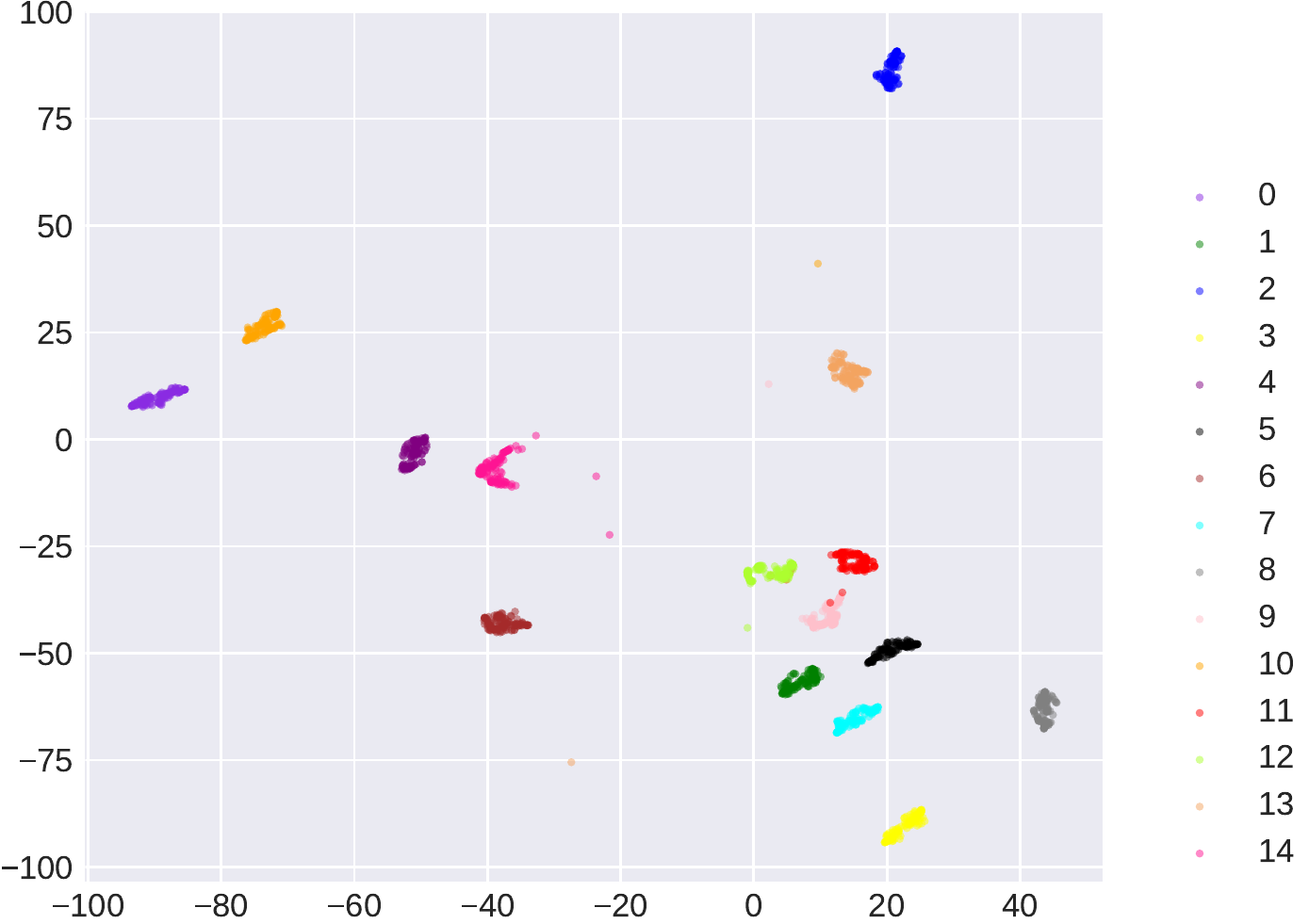}
    }
    % \subfigure[single(ours)]{
        % \includegraphics[scale=0.375]{fig/cluster-level.pdf}
    % }
    \vspace{-0.3cm}
    \caption{IND \& OOD intents visualazation of DeepAligned and E2E method, we select 9 IND intents and 6 OOD intents in GID-MD-40\% (index 0-8 denotes IND intents, index 9-14 denotes OOD intents) }
    \label{fig: visualazation}
    %  \vspace{-0.5cm}
\end{figure*}

\subsubsection{Estimate the Number of Cluster K}
\label{k}
All the results we showed so far assume that the number of OOD classes is pre-defined. However, in real-world applications, this often needs to be estimated automatically. Table \ref{tab:k} shows the results using the same  automatic K-value estimation strategy \footnote{Here we use the same estimation algorithm as \citet{Zhang2021DiscoveringNI}. We leave the details in Appendix \ref{appendix_k}.}. We find that our method both achieves the best performance under the fixed or auto K settings. It should be noted that no matter the end-to-end methods or the pipeline methods, the performance drops significantly when the number of OOD classes is unknown. Therefore, how to estimate an accurate K value and how to design a more robust GID method is a great challenge.

\subsubsection{Visualization}

\begin{table}[]
\centering
\resizebox{0.45\textwidth}{!}{%
\begin{tabular}{l|c|c|c|c}
\hline
\multicolumn{1}{c|}{} & OOD ACC     & OOD F1     & ALL ACC    & K  \\ \hline
DeepAligned             & 87.55   & 87.14   & 93.70  & 60 \\
DeepAligned-Mix       & 82.70   & 84.65   & 88.12  & 60 \\
End-to-End       & 91.92   & 92.46   & 95.78  & 60   \\ \hline
DeepAligned           & 72.89   & 66.75   & 87.91  & 47 \\
DeepAligned-Mix     & 69.56   & 62.32   & 85.29   & 47 \\
End-to-End       & 74.89   & 67.23   & 88.58  & 47   \\ \hline
\end{tabular}%
}
% \vspace{-0.2cm}
\caption{Estimate the number of OOD clusters. K=47 is the estimated number compared to original 60.}
\vspace{-0.5cm}
\label{tab:k}
\end{table}

To further visually compare the performance of end-to-end and pipeline methods in classifying IND and clustering OOD, we performed a visualization of IND \& OOD intent representations for E2E and DeepAligned, as shown in Fig \ref{fig: visualazation}. Comparing E2E to DeepAligned, we can observe DeepAligned gets some mixed OOD clusters (see greenyellow and red dots in Fig a) while E2E method successfully separates them. We also find that many OOD intents in DeepAligned that cannot be clustered into single cluster, but are scattered into multiple clusters (see deeppink dots in Fig a), but E2E method can form compact clusters for them. We argue this is because the pipeline method introduces serious error propagation in the OOD clustering stage; while the E2E method jointly learns OOD cluster assignments and classification of IND \& OOD, which helps to get clear cluster boundary.

%many OOD intents in DeepAligned that cannot be clustered into single cluster, but are scattered into multiple clusters, but End-to-End method can form clear-bounded intent clusters for both IND and OOD. We argue this is because the pipeline method DeepAligned introduces serious error propagation in the OOD clustering stage; while the End-to-End method jointly learns OOD cluster assignments and classification of IND \& OOD, the interaction between IND \& OOD improve the accuracy for OOD pseudo labels.

\subsubsection{Noise of IND}
In the general GID setting, we assume that the IND and OOD categories do not overlap, however the OOD data collected in practical application scenarios may have some IND noise due to the error propagation of OOD detection. We analyze the performance changes of each GID method when mixing different proportions of IND noise in OOD data, as shown in Fig \ref{fig:IND_noise}. The results show that our E2E method still significantly outperforms the pipeline baseline under IND noise scenarios. The performance of IND classification and OOD clustering for all methods decrease significantly, and the IND performance decrease is more significant for DeepAligned and E2E. We argue that this is due to the inclusion of a small amount of IND data in the OOD data, which causes these IND data to be incorrectly labeled, and severely impairs the performance of IND classification, making it difficult to form clear IND class boundaries. We also found that when the IND noise ratio reached 15\%, the OOD clustering performance of the E2E method was worse than DeepAligned. We argue that this is because the E2E method jointly learns to classify IND intents and discover OOD intents, which needs to leverage IND prior knowledge to enhance OOD clustering. However, When there is more IND noise to be mixed with OOD data, it will affect the effectiveness of the knowledge interaction between IND and OOD. In practical applications, when the performance of OOD detection is improved, this IND noise problem can be relieved naturally, which is not within the scope of this papar.

\section{Challenges}
\label{challenges}
Based on the above analysis, we summarize the current challenges faced by the GID task: 

\textbf{Fine-grained OOD data} When OOD intents are fine-grained like GID-SD, the OOD performance of existing GID methods decreases significantly. We argue fine-grained OOD intents make it hard to construct clear boundary while clustering. 
% This is mainly because current methods do not specifically target fine-grained OOD intent. This is a potential challenge. 

\textbf{Cross-domain transfer} When IND and OOD intent types are from different distant domains, the knowledge learned from IND is hard to transfer to OOD due to the semantic gap in different domains. 
% Therefore, cross-domain transfer is an important challenge for GID tasks. 

\textbf{OOD noise} 
% In the general setting, we assume that the number of OOD categories is known, and the OOD data must belong to a certain intent type. 
OOD data collected in practical applications are usually noisy, and there may be some OOD samples that do not belong to a certain intent type. The performance of each GID method degrades when trained with these noisy OOD data.
% Therefore, OOD noise is a very practical challenge. 

\textbf{imbalanced OOD data} The OOD data in real-world scenarios is often class-imbalanced, and our analysis in section \ref{imbalanced} proves that the performance of current methods drops significantly under imbalanced data, especially end-to-end methods. 
% Therefore, class-imbalanced GID is also a potential challenge. 

\textbf{Inaccurate estimation of the number of OOD categories} Most previous work assume the number of OOD categories is known. However, in practical applications, we usually need to estimate the number of categories K, which is often inaccurate. We propose a preliminary analysis in Section \ref{k} which shows significant performance drop when the estimation is not totally accurate.
% Therefore, a key challenge is how to ensure the robustness of the GID model under inaccurate K values. 

\section{Related Work}
% OOD相关：带一下CV
% 增量学习：强调标签集合给定，通常是有标注的

\textbf{OOD Detection} aims to know when a query falls outside the range of pre-defined supported intents \cite{Zeng2021ModelingDR, Lin2019DeepUI, Xu2020ADG,Zeng2021AdversarialSL,Wu2022RevisitOF} to avoid performing wrong operation. OOD detection has attracted more and more attention in recent years, so various similar names are derived, such as anomaly detection, open world classification \cite{shu2021odist}, open-world learning \cite{xu2019open}, open intent classification\cite{zhang2021deep} and so on. However, all of them are essentially to distinguish whether a query belongs to IND or OOD intents, without further discovering new semantic categories from unsupervised OOD data. 

\textbf{OOD Discovery} aims to discover new intent concepts from unlabeled OOD data and form OOD intent clusters \cite{Lin2020DiscoveringNI,Zhang2021DiscoveringNI,mou-etal-2022-disentangled}, which focuses more on how to cluster OOD data, while ignoring the fusion of IND and OOD, which makes the model only recognize OOD intents. For example, \cite{Zhang2021DiscoveringNI} design an iterative clustering algorithm DeepAligned, which iteratively learns intent representations then cluster assignments. \textbf{Open Intent Extraction} also aims to extract unknown intents from unlabelled user queries \cite{vedula2020open}, and is a completely unsupervised task. However, in terms of method, open intent extraction is more about extracting intent names through sequence annotation methods. In contrast, GID aims to train a network that can simultaneously classify a set of labeled IND intent classes while discovering and recognizing unlabeled OOD intents.

\textbf{Incremental/Continual Learning} There is currently some work on extending closed-set classifier to new classes in the open world incrementally, such as \cite{xu2019open}.
But all these works follow a traditional incremental learning setting, which requires new category data with labels. In practical applications, we can only obtain these unlabeled OOD data from the dialogue system logs, and these data are often updated continuously, and human annotation of these data is very labor-intensive. Therefore, we propose a more human-free task GID, which aims to automatically discover new categories from the unlabeled OOD data, and further expand the recognition scope of the existing IND intent classifier incrementally.

\textbf{Zero-shot Intent Detection} Zero-shot intent detection \cite{yan2020unknown, siddique2021generalized} assumes that no target domain data is available during training, but the category and category descriptions from target domain are given, but in practical applications we often have access to a large amount of unlabeled dialogue logs, and we need to consider how to discover new intent categories from them for system development.

\section{Conclusion}

In this paper, we introduce a new task, Generalized Intent Discovery (GID), which aims to extend an IND intent classifier to an open-world intent set. Then we provide three public datasets for different application scenarios and establish a benchmark for the GID task. We also propose extensive baselines of two frameworks, pipeline-based and end-to-end for future work. Further, We conduct exhaustive experiments and qualitative analysis to comprehend key challenges and provide new guidance for future GID research.

\section*{Acknowledgements}
We thank all anonymous reviewers for their helpful comments and suggestions. This work was partially supported by MoE-CMCC "Artifical Intelligence" Project No. MCM20190701, National Key R\&D Program of China No. 2019YFF0303300 and Subject II No. 2019YFF0303302, DOCOMO Beijing Communications Laboratories Co., Ltd.

\section*{Broader Impact}
Task-oriented dialogue systems have demonstrated remarkable performance in a wide range of applications, and have significant positive impact on human production mode and lifeway. Intent classification is an important component of
task-oriented dialogue system. Existing intent classification models can only identify a limited number of predefined in-domain (IND) intents, however, out-of-domain (OOD) or unknown intents will appear continually when the dialogue system is deployed online. If we can group these OOD samples into different clusters, we can discover new intents, guide future development of the system, and expand the classification capabilities of the system. We note that OOD intent detection and OOD intent discovery tasks have been widely studied recently. The former focuses on identifying whether a sample is IND or OOD, while the latter focuses on how to cluster OOD data. The generalized intent discovery (GID) task proposed in this paper focuses on an incremental setting, that is simultaneously classifying a set of labeled IND intent classes while discovering and recognizing new unlabeled OOD types incrementally. GID aims to provide the model with the ability to automatically learning according to known knowledge in the open world, which is a new attempt for scalable dialogue system and open world learning.

\bibliography{anthology,custom}

\appendix

\section{Appendix}
\label{sec:appendix}

\subsection{Original Intent Dataset Statistics}
\label{statistics}

\begin{figure*}[t]
    \centering
    \resizebox{0.9\linewidth}{!}{
    \includegraphics{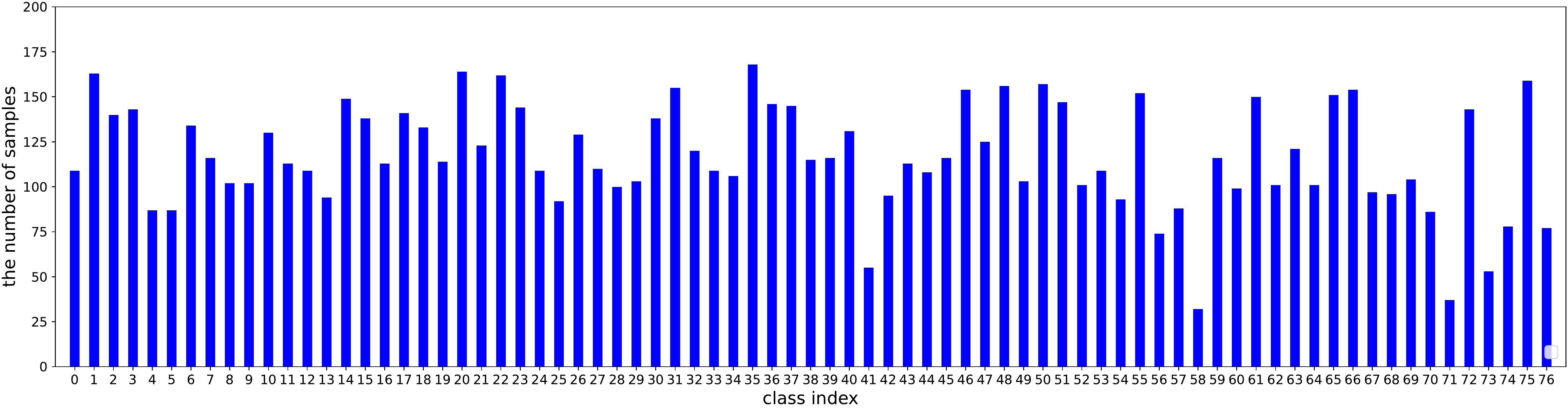}}
    % \vspace{-0.2cm}
    \caption{The number of samples for each class in Banking dataset.}
    \label{fig:banking}
    %  \vspace{-0.5cm}
\end{figure*}

We show the detailed statistics of CLINC and BANKING datasets in Table \ref{tab:dataset2}. Banking is class-imbalanced, and the number of samples for each class is shown in Fig \ref{fig:banking}. The three GID datasets GID-SD GID-MD and GID-CD we constructed in this paper, all maintain the same train/dev/test split as the original dataset. Table \ref{tab:GID-md} shows the number of intents divided into IND and OOD per domain for GID-MD-40\%. Since CLINC and BANKING are open source datasets, there is no license problem.

\begin{table*}[h]
		\centering
		\resizebox{0.9\textwidth}{!}{
		\begin{tabular}{ ccccccc }
			\toprule
			Dataset & Classes & Training & Validation & Test & Vocabulary & Length (max / mean) \\
			\midrule
			CLINC & 150  & 18,000 & 2,250 & 2,250 & 7,283 & 28 / 8.31 \\
			BANKING & 77  & 9,003 & 1,000 & 3,080 & 5,028 & 79 / 11.91 \\ 
			\bottomrule
		\end{tabular}}
		\caption{Statistics of CLINC and BANKING datasets.}
% 		\vspace{-0.7cm}
		\label{tab:dataset2}
\end{table*}

\begin{table}[h]
		\centering
		\resizebox{.48\textwidth}{!}{
		\begin{tabular}{ ccccccc }
			\toprule
			Domains & IND intents & OOD intents\\
			\midrule
			banking & 10  & 5 \\
			credit\_cards & 8  & 7  \\
			kitchen\_and\_dining & 9  & 6  \\
			home & 6  & 9  \\
			work & 10  & 5  \\
			utility & 8  & 7  \\
			travel & 9  & 6  \\
			auto\_and\_commute & 10  & 5  \\
			small\_talk & 11  & 4  \\
			meta & 9  & 6  \\
			\bottomrule
		\end{tabular}
		}
		\caption{The number of intents divided into IND and OOD per domain for GID-MD-40\%}
% 		\vspace{-0.35cm}
		\label{tab:GID-md}
\end{table}

\subsection{GID-imbalanced}
\label{imbal}
\begin{figure}
    \centering
    \resizebox{.48\textwidth}{!}{
    \includegraphics{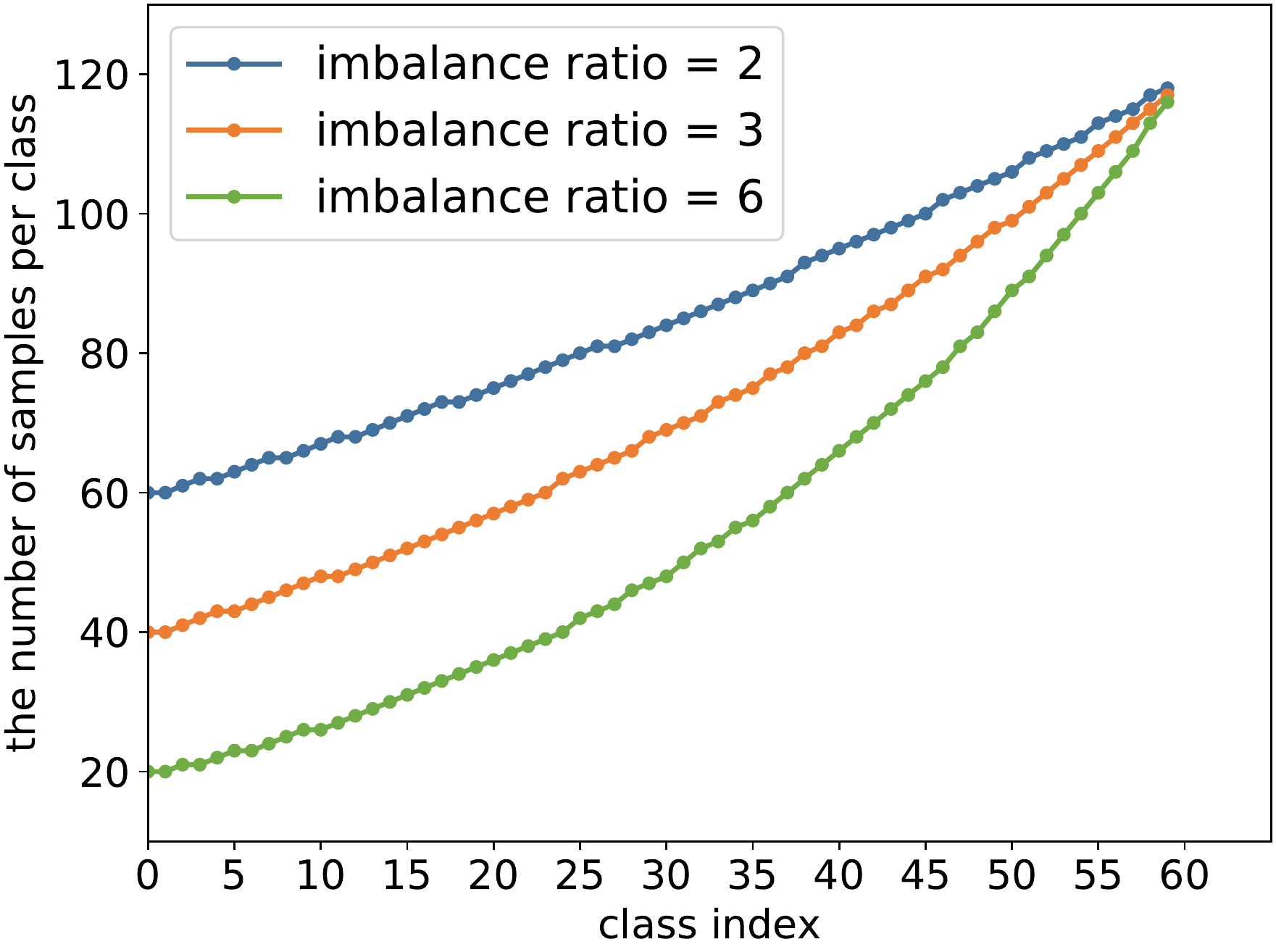}}
    \caption{The distribution of the number of samples for GID-imbalance}
    \label{fig:longtail}
    % \vspace{-0.3cm}
\end{figure}

For our imbalanced dataset GID-imbalance, we show the distribution of the number of samples per OOD category under the influence of different imbalance ratio($\rho=2,3,6$) in Figure\ref{fig:longtail}. The larger the imbalance ratio, the more significant the class imbalance degree of the corresponding imbalanced dataset. 

\subsection{Implementation Details}

For a fair comparison of the various methods, we use the pre-trained BERT model (bert-base-uncased \footnote{https://github.com/google-research/bert}, with 12-layer transformer) as our network backbone, and add a pooling layer to get intent representation(dimension=768). Moreover, we freeze all but
the last transformer layer parameters to achieve better performance with BERT backbone, and speed up the training procedure as suggested in \cite{Zhang2021DiscoveringNI}. Firstly, we use labeled IND data to pretrain BERT model. For pipeline method(k-means and DeepAligned), we use the official implementation and hyperparameters offered by \cite{Zhang2021DiscoveringNI} to realize it, and the batch size is 512 and learning rate is 5e-5 for joint classification stage. For DeepAligned-Mix, the training batch size is 512 and the learning rate is 5e-5. For end-to-end method, IND head and OOD head are two symmetrical MLPs (input dimension is 768 and output dimension is the number of categories for IND/OOD), and we select 
$tanh$ as activation function as previous work. We use SGD with momentum as optimizer, with linear warm-up and cosine annealing ($lr_{base}$ = 0.4, $lr_{min}$ = 0.01), and weight decay $10^{-4}$. The batch size is always set to 512 for all experiments. Notably, We use dropout \cite{Gao2021SimCSESC} to construct augmented examples and the dropout value is fixed at 0.5. For what concerns pseudo-labeling, we use the implementation of the Sinkhorn-Knopp algorithm provided by \cite{caron2020unsupervised} and we inherit all the hyperparameters from \cite{caron2020unsupervised}, e.g. $n\_iter$ = 3 and $\epsilon$ = 0.05. We use the SC value of the validation data to select the best checkpoints. All experiments use a single Tesla T4 GPU(16 GB of memory). Table\ref{tab:implementation_detail} shows the comparison of the epoch and training time required for the convergence of the End-to-End method and DeepAligned. We can see that the E2E method takes fewer epochs to converge than the DeepAligned method. This is because the DeepAligned method first performs OOD clustering, and then uses the obtained OOD pseudo-labels and IND ground-truth labels for joint classification, which will lead to The OOD pseudo-labels have serious errors, and these label errors cannot be corrected in classification process, resulting in difficulty in model convergence. In addition, we can also see that the E2E method only increases the time required for each epoch by 1.8s compared to the classification stage of DeepAligned, which indicates the efficiency of the SK algorithm.

\begin{table}[h]
		\centering
		\resizebox{.48\textwidth}{!}{
		\begin{tabular}{ ccccccc }
			\toprule
			Method & training epoch & training time\\
			\midrule
			End-to-End & 51  & 30s/epoch \\
			DeepAligned(two-stages) &   &  \\ 
			- clustering & 67  & 27.6s/epoch \\
		    - classification & 91  & 28.2s/epoch \\ 
			\bottomrule
		\end{tabular}}
		\caption{Comparison of training efficiency between pipeline and End-to-End methods. We use the same hardware.}
% 		\vspace{-0.7cm}
		\label{tab:implementation_detail}
\end{table}

% \begin{figure}
%     \centering
%     \resizebox{.48\textwidth}{!}{
%     \includegraphics{figures/dataset_construction.pdf}}
%     \caption{The illustration of construction strategies for three benchmark GID datasets.}
%     \label{fig:construction}
%     \vspace{-0.3cm}
% \end{figure}

\subsection{Estimate K}
\label{appendix_k}
\begin{figure*}[t]
    \centering
    \resizebox{0.9\linewidth}{!}{
    \includegraphics{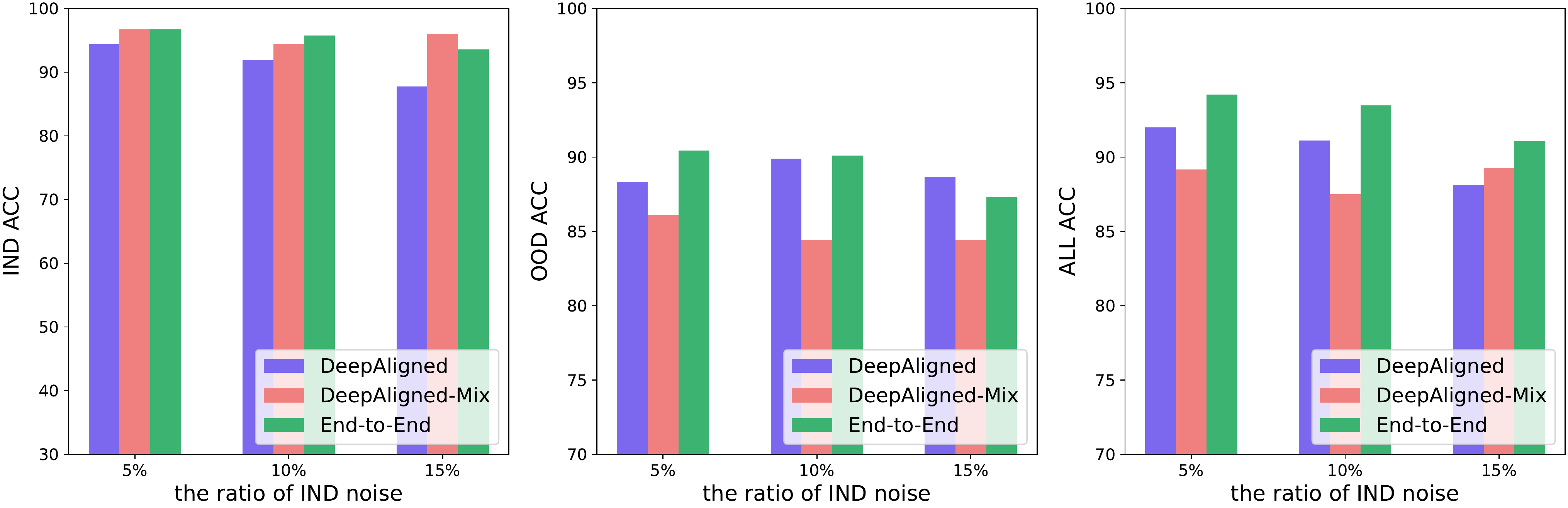}}
    % \vspace{-0.2cm}
    \caption{The impact of adding different ratios of IND noise samples to the OOD training data on the performance of each GID model.}
    \label{fig:IND_noise}
    %  \vspace{-0.65cm}
\end{figure*}

Since we may not know the exact number of OOD clusters, we use the following K estimation method \cite{Zhang2021DiscoveringNI} to determine the number of clusters K before clustering. The method estimates K with the aid of the well-initialized intent features. We assign a big $K^{\prime}$ as the number of clusters at first. As a good feature initialization is helpful for partition-based methods (e.g., k-means), we use the well pre-trained model to extract intent features. Then, we perform k-means with the extracted features. We suppose that real clusters tend to be dense even with $K^{\prime}$, and the size of more confident clusters is larger than some threshold $t$. Therefore, we drop the low confidence cluster whose size is smaller than $t$, and calculate K with:
\begin{align}
    K=\sum_{i=1}^{K^{\prime}} \delta\left(\left|S_{i}\right|>=t\right)
\end{align}
where $\left|S_{i}\right|$ is the size of the $i^{t h}$ produced cluster, and $\delta(\cdot)$ is an indicator function. It outputs 1 if condition is satisfied, and outputs 0 if not. Notably, we assign the threshold $t$ as the expected cluster mean size $\frac{N}{K^{\prime}}$ in this formula.

\subsection{Effect of IND Data}
\begin{figure}[t]
    \flushleft
    \resizebox{0.98\linewidth}{!}{
    \includegraphics{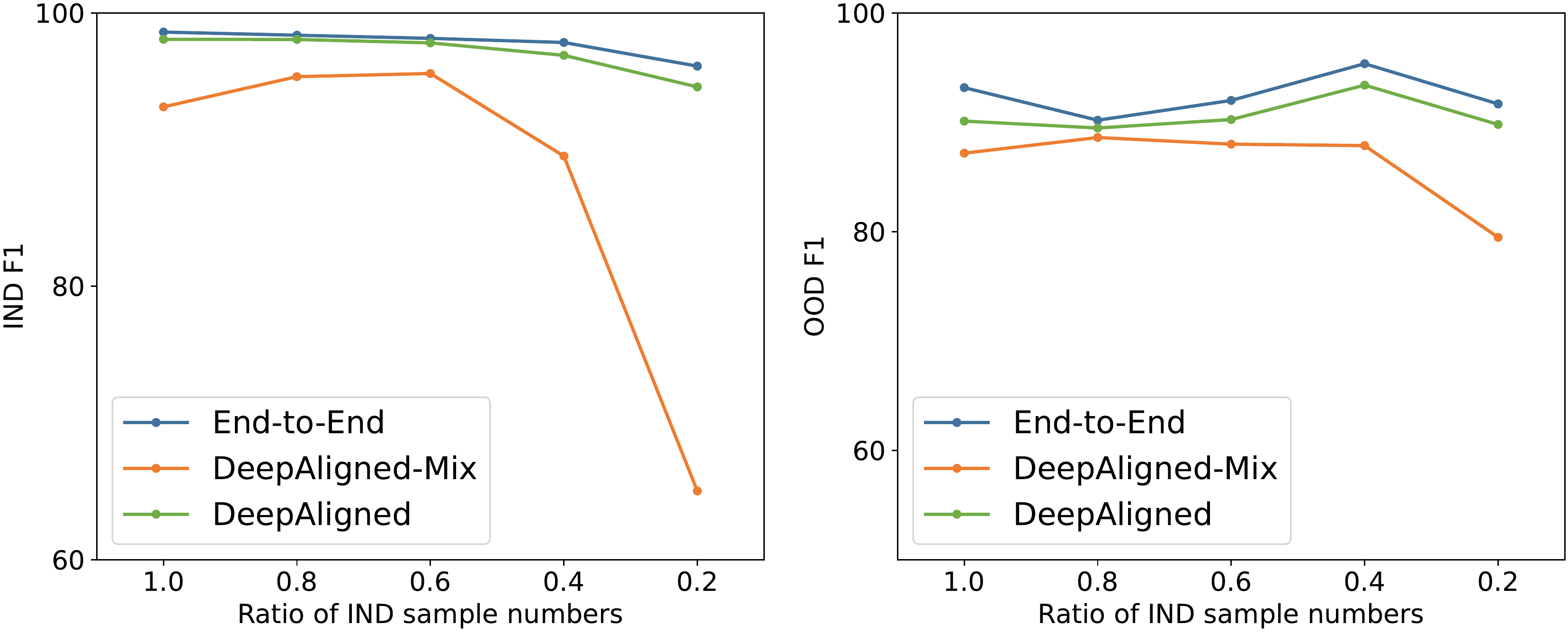}}
    % \vspace{-0.2cm}
    \caption{ Effect of IND data for GID. The left subfig denotes IND F1 and the right subfig denotes OOD F1.}
    \label{Effect of IND Data}
    %  \vspace{-0.65cm}
\end{figure}

We analyze the impact of the number of samples per IND class on the performance of each model. Fig \ref{Effect of IND Data} shows the trend of model performance as the number of IND samples for each class decreases. Overall, the performance of our end-to-end method is much better than the baselines. Moreover, with the decrease of the amount of in-domain data, all methods show varying degrees of performance fluctuation. We observe the changes of IND F1 and OOD F1, and find IND F1 generally shows a downward trend, especially for DeepAligned-Mix. We believe that this is because the number of IND samples in each category is reduced, resulting in the biased joint classification of IND\&OOD towards the OOD category. DeepAligned-Mix learns both IND and OOD by clustering, which will lead to inaccurate pseudo-labels obtained by IND, further degrading the performance. As for OOD F1, due to the reduced number of IND samples, the model can learn less IND prior knowledge, thus affecting the performance of OOD. Therefore, GID in the small labeled IND scenario is also a challenge worthy of attention.

\subsection{Silhouette Coefficient (SC)}
\label{sc}
Following \citet{Zhang2021DiscoveringNI}, we use the cluster validity index (CVI) to evaluate the quality of clusters obtained during each training epoch after clustering. Specifically, we adopt an unsupervised metric Silhouette Coefficient \cite{rousseeuw1987silhouettes} for evaluation:
\begin{align}
    S C=\frac{1}{N} \sum_{i=1}^{N} \frac{b\left(\boldsymbol{I}_{i}\right)-a\left(\boldsymbol{I}_{i}\right)}{\max \left\{a\left(\boldsymbol{I}_{i}\right), b\left(\boldsymbol{I}_{i}\right)\right\}}
\end{align}
where $a\left(\boldsymbol{I}_{i}\right)$ is the average distance between $\boldsymbol{I}_{i}$ and all other samples in the $i$-th cluster, which indicates the intra-class compactness. $b\left(\boldsymbol{I}_{i}\right)$ is the smallest distance between $\boldsymbol{I}_{i}$ and all samples not in the $i$-th cluster, which indicates the inter-class separation. The range of SC is between -1 and 1, and the higher score means the better clustering results.

\end{document}